# Predicting Metabolic Dysfunction-Associated Steatotic Liver Disease using Machine Learning Methods: A Retrospective Cohort Study

Mary E An, MS[1], Paul M Griffin, PhD[1], Jonathan G Stine, MD[2], Balakrishnan S Ramakrishna, MD[3], Soundar RT Kumara, PhD[1]

[1]Harold and Marcus Inge Department of Industrial and Manufacturing Engineering, The Pennsylvania State University, University Park, PA, US

[2]Penn State College of Medicine, The Pennsylvania State University, Hershey, PA, US
[3]Department of Gastroenterology, SRM Institutes for Medical Science, Chennai, Tamil Nadu, IN

**Corresponding Author:** Soundar RT Kumara, u1o@psu.edu

## Abstract

**Background:** Metabolic dysfunction-associated steatotic liver disease (MASLD) affects 30%-40% of US adults and is the most common chronic liver disease. Although often asymptomatic, progression can lead to cirrhosis.
**Objective:** The objective of the study was to develop and evaluate an electronic health record (EHR) based prediction model to support early detection of MASLD in primary care settings.
**Methods:** We evaluated LASSO logistic regression, random forest, XGBoost, and a neural network model for MASLD prediction using clinical feature subsets from a large EHR database, including the top 10 ranked features. To reduce disparities in true positive rates across racial and ethnic subgroups, we applied an equal opportunity postprocessing method in a prediction model called MASLD EHR Static Risk Prediction (MASER).
**Results:** This retrospective cohort study included 59,492 participants in the training data, 24,198 in the validating data, and 25,188 in the testing data. The LASSO logistic regression model with the top 10 features was selected for its interpretability and comparable performance. Before fairness adjustment, the model achieved AUROC of 0.84, accuracy of 78%, sensitivity of 72%, specificity of 79%, and F1-score of 0.617. After equal opportunity postprocessing, accuracy modestly increased to 81% and specificity to 94%, while sensitivity decreased to 41% and F1-score to 0.515, reflecting the fairness trade-off.
**Conclusions:** MASER achieved competitive performance for MASLD prediction, comparable to previously reported ensemble and tree-based models, while using a limited and routinely collected feature set and a diverse study population. The model is designed to support early detection and potential integration into primary care workflows. MASER demonstrates EHR-ready MASLD prediction with fairness adjustments, supporting future primary care implementation pending prospective validation.

## Introduction

Metabolic dysfunction-associated steatotic liver disease (MASLD) refers to a range of conditions caused by excess fat in the liver of a person who drinks little to no alcohol. Prevalence of MASLD is approximately 33% of US adults,[1] and the total economic burden of the disease in direct costs was estimated to be $103 billion in 2016.[2] It is projected that economic burden could increase to over $1 trillion in 2026 and prevalence could increase to 41% by 2050.[2,3]

To properly diagnose MASLD, liver biopsy remains the gold standard. However, liver biopsies are invasive, leading to patient burden and anxiety, and expensive, and are therefore avoided when possible.[1,4] Instead, it is common to diagnose a patient with MASLD through a combination of blood and imaging tests.[5–8] If uncorrected, MASLD can progress to MASH with or without liver fibrosis. In the MASH population, liver fibrosis stage drives mortality; and if progression to cirrhosis and liver cancer happens, liver transplantation may be necessary.

There are two regulatory agency-approved (conditional/accelerated) medications, resmetirom and semaglutide. However, they are reserved for F2-F3 disease, and their efficacy remains somewhat limited with even greater rates of real-world discontinuations. Early detection of MASLD to initiate healthy lifestyle changes or MASH with fibrosis to initiate medications is important to improve patient outcomes. Examples include the Fatty Liver Index (FLI), the Zhejiang University (ZJU) Index, and the Hepatic Steatosis Index (HSI). The FLI is based on data from the Dionysos Nutrition and Liver Study, done with a non-Hispanic White population.[9] The FLI study uses BMI, waist circumference, triglycerides, and gamma-glutamyl transferase to determine if a patient is at risk for fatty liver.[10] The HSI is based on data from a Korean population, and the risk index is based on ALT/AST ratio, BMI, and T2DM.[11] The ZJU index used data from a Chinese population and includes the factors: BMI, fasting plasma glucose, triglycerides, and ALT/AST ratio.[12] Since all these indices were developed based on study populations that tend to be of the same race and/or ethnicity, it is not apparent whether these are directly applicable to more diverse populations. Additionally, waist circumference, seen in the FLI, is typically not included as a routine measurement that physicians collect, which also supports the need for a new index.[13,14]

According to a scoping review completed by Talens et al (2021) the Hispanic population has a higher prevalence of MASLD (possibly due to the presence of the PNPLA3 gene) while the non-Hispanic Black population has the lowest prevalence.[15] Additionally, although the Hispanic population has the highest prevalence, they suffer poorer outcomes compared to the non-Hispanic White population. It was also found that the non-Hispanic White population has fewer unfavorable outcomes compared to other race/ethnicity populations.[15] Rich et al (2018) confirm these findings but urge researchers to discover why these disparities exist.[16] They note diabetes and obesity are common to both the Hispanic population and the non-Hispanic Black population.[16] It is possible the Hispanic population has a higher prevalence and worse outcomes due to other factors such as genetics, limited access to healthcare, health insurance coverage, etc.[15,16] These findings highlight the importance of subgroup analysis to enhance model transparency.

There has been an increase in the utilization of machine learning tools to predict the presence of diseases such as MASLD due to improved operational efficiency and performance.[17] However, these tools are rarely utilized in clinical practice primarily due to lack of trust in the developed models and barriers to implementation in clinical workflows.[18] These concerns stem from conventional machine learning workflows, in which all available variables are often incorporated without sufficient clinical curation or consideration of real-world feasibility. We aimed to address these concerns by developing a prediction model for whether a patient currently has any stage of MASLD. We focused on (1) the diverse United States (US) population to develop an equal opportunity (fair) MASLD prediction model that has similar model performance across different races and ethnicities, (2) a limited feature set which only includes features collected at routine primary care visits, and (3) a new tool, MASLD Static Electronic Health Record (EHR) Risk Prediction (MASER), to enhance early detection and appropriate referral for treatment.

## Methods

This study was conducted and reported in accordance with the Strengthening the Reporting of Observational Studies in Epidemiology [19] statement and the Transparent Reporting of a multivariable prediction model for Individual Prognosis Or Diagnosis [20] guidelines.

### *Ethical Considerations*

Data used in this study came from TriNetX, a global health research network that provides "de-identified data based on the standard defined in Section §164.514(a) of the HIPAA Privacy Rule. The process by which data sets are de-identified is attested to through a formal determination by a qualified expert as defined in Section §164.514(b)(1) of the HIPAA Privacy Rule [21]." The data is continuously updated. It is also cleaned and transformed to comply with TriNetX's common data model to ensure data quality and usability by the researcher. Because this data is de-identified by TriNetX, it was determined by the Human Subjects Protection Office at Penn State that TriNetX research does not meet the definition of human subject research and is therefore exempt from IRB approval.

### *Study Population*

All data for this study was collected on 05/01/2024 from the TriNetX Research Network and includes data from October 1, 2001 to April 30, 2024. The data includes a total of 7,824,804 patients separated into two cohorts: patients with MASLD and patients without MASLD. Inclusion criteria for the MASLD cohort were (1) patient must be 18 years or older, (2) diagnosed with fatty liver (ICD-10-CM K76.0) or NASH (ICD-10-CM K75.81). Inclusion criteria for the non-MASLD cohort were (1) patient must be 18 years or older, (2) general adult medical examination recorded in the medical records (ICD-10-CM Z00.0); and exclusion criteria for this cohort was no fatty liver or NASH diagnosis anywhere in their medical records. Exclusion criteria for both cohorts can be divided into three main sections: alcohol use, other causes of liver disease, and secondary causes of steatosis [22,23].

Finally, an index event was defined and to be used during the propensity score matching process. After exporting data from TriNetX, the data was filtered using ICD-9-CM codes for the same inclusion and exclusion criteria as described previously in this section. The data selection process is captured in Figure 1, and the inclusion and exclusion criteria with their ICD-10-CM codes can be found in supplementary multimedia appendix (SMA) Tables S1-S3. The ICD-9-CM codes corresponding to the ICD-10-CM codes can be found in SMA Extended Data Table.

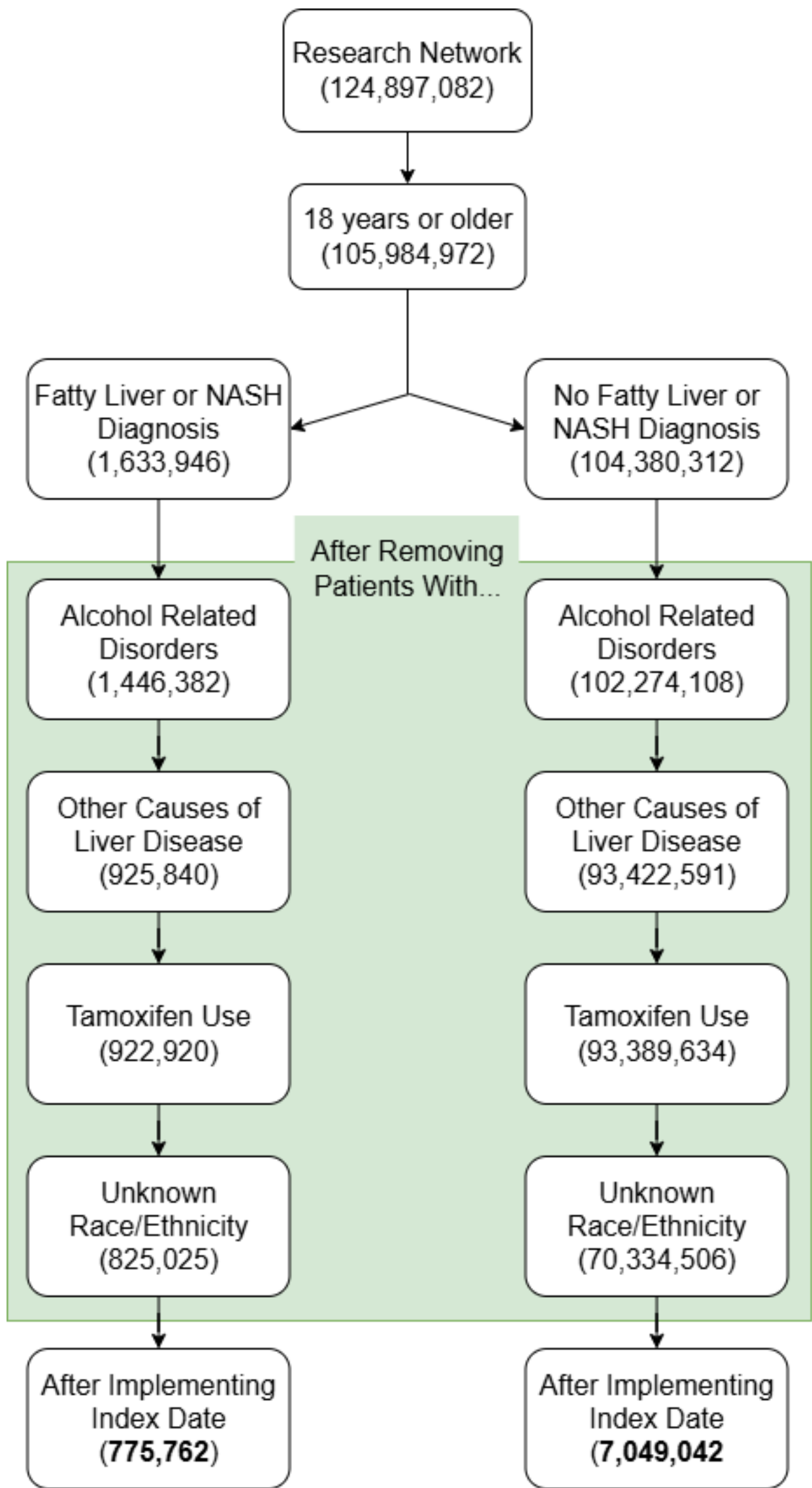

*Figure 1. Inclusion and Exclusion Criteria with ICD-10-CM codes and subsequently with ICD-9-CM codes. Abbreviations: ICD-9-CM, international classification of diseases, 9th revision, clinical modification; ICD-10-CM, international classification and diseases, 10th revision, clinical modification; NASH, non-alcoholic steatohepatitis.*

***Data Preprocessing***

TriNetX provides data in multiple tables with the patient ID as the key. For our analysis, four tables were used: patient, diagnosis, lab result, and vital signs. Data processing was performed using PySpark (Apache Spark version 3.2.0) in Python within a Jupyter Notebook environment.

To avoid overfitting the model, the data was divided at random by patient ID into training, validating, and testing datasets. After confirming that the MASLD dataset only contained patients with an MASLD diagnosis and similarly that the non-MASLD dataset did not contain any patients with an MASLD diagnosis, each cohort of data was randomly split with the following distributions: the training data was 70% of each dataset, the validating data 15%, and the testing data 15%.

Preprocessing began with the patient table since it contains demographic information. First, age categories were created from the age column using the following bins[24]: 18-34, 35-49, 50-64, 65 or greater. Next, a race/ethnicity column was created by condensing race and ethnicity into five main categories of non-Hispanic White, non-Hispanic Black, non-Hispanic Asian, non-Hispanic other, and Hispanic. With the diagnosis table, binary variables were added to each patient in the patient table to indicate whether that patient had ever been diagnosed with type 2 diabetes mellitus (T2DM), hypertension, or had been a smoker.

***Selection and Preparation of Demographic and Clinical Features***

Features were selected based on the literature review and on what primary care physicians collect at regular visits. Primary care physicians typically collect a basic metabolic panel or a comprehensive metabolic panel; therefore, features include sex, age, race/ethnicity, high-density lipoprotein cholesterol (HDL), low-density lipoprotein cholesterol (LDL), alanine transaminase (ALT), aspartate aminotransferase (AST), alkaline phosphatase (ALP), triglycerides (TG), body mass index (BMI), blood urea nitrogen (BUN), creatinine (Cr), bilirubin (BIL), albumin (ALB), total protein (TP), fasting plasma glucose (FPG), smoker or non-smoker. To ensure data quality and prevent the influence of extreme values, we capped the laboratory and vital sign features within clinically plausible ranges. The ICD-10-CM codes, ICD-9-Codes, LOINC codes, and capping thresholds are detailed in SMA Table S4.

***Propensity Score Matching***

To minimize confounding, 1:1 propensity score matching was performed on the training data. MASLD and non-MASLD patients were matched on sex and age. The matching was done by identifying the number of patients in the MASLD group that has a certain combination of matching features (sex and age) and then sampling the same number of patients from the non-MASLD group who also has the same combination of matching features. This is done until all combinations of matching features are exhausted.

The index event used for the MASLD cohort was the date of their first MASLD diagnosis while the index event for the non-MASLD cohort was the first date of a general adult medical examination encounter. The date of the index event for each patient was identified, and all tables (vitals, labs, and diagnoses) were filtered to only include data from on or up to one year before the index event. This ensured that all data was current for each patient. Finally, the median value for each feature of interest was used in the final analysis.

***Statistical Analysis***

Statistical analysis was completed in Python (version 3.9.10). Categorical features were presented as counts and percentages while continuous features were summarized as mean and standard deviation. For each categorical feature, a chi-square test was performed to test for differences between the MASLD and non-MASLD distributions before and after matching. For each continuous feature, a Mann-Whitney U test was used. These statistical tests provide a baseline understanding of how each cohort differs for each feature with $p<0.05$ indicating a statistically significant difference.

### *Machine Learning Workflow and Validation Strategy*

Four different modeling methods were considered: LASSO logistic regression, random forest, XGBoost, and a fully connected neural network. LASSO logistic regression is a technique that combines the least absolute shrinkage and selection operator (LASSO), a regularization method, with a logistic regression model. This method shrinks coefficients of less important predictors preventing overfitting and improving model interpretability. Random forest uses bagging and decision trees to make initial predictions then chooses the prediction of majority as the final prediction. This method is called random forest because the decision trees are generated using random feature selection. Extreme Gradient Boosting (XGBoost) sequentially builds decision trees while correcting the errors of the previous trees. It then combines the trees by adding their predictions together. The neural network is inspired by the human brain and can act autonomously; therefore, input by the user is simply for fine tuning parameters to achieve a better result. A deep neural network is simply a neural network with more than two hidden layers making it more complex. Neural networks are attractive as there are no assumptions with respect to the data that need to be met and typically have a high accuracy.

Continuous features were standardized to zero mean and unit variance for the logistic regression and the neural network models, in accordance with their algorithmic assumptions. The random forest and XGBoost models were trained using unscaled features as these methods are invariant to monotonic feature transformations. To create a cleaner decision boundary in the logistic regression training phase, the SMOTETomek resampling method from the Imbalanced-learn package[25] was applied. No resampling was applied during validation or testing. Hyperparameter tuning was performed using the training set via grid or randomized search, depending on the algorithm. Model performance was monitored on a separate validation set to assess signs of underfitting or overfitting and guide any further parameter adjustments. Final model performance was evaluated on the held-out testing dataset.

After an initial model for each method was trained using all features on the training set, SHAP (Shapley Additive exPlanations) analysis was conducted to determine feature importance. SHAP values are used for machine learning explainability by quantifying the contribution of each feature on the predicted output for individual data points.[26] After SHAP analysis, three additional models were developed for each method using only the top 10, top 5, and top 3 SHAP features, respectively. The final hyperparameters for all 16 models are included in the supplementary multimedia appendix.

In the validation and testing datasets, the proportion of MASLD to non-MASLD patients was adjusted to 1:3 to reflect the estimated prevalence of MASLD in the US population. This stratification was maintained to ensure that model evaluation occurred under realistic epidemiologic conditions.

### *Evaluation Strategy*

To gain a complete understanding of how the models perform, several evaluation metrics were deployed. AUROC, the area under the receiver operating characteristic (ROC) curve, is a metric used to evaluate how well a model discriminates between positive and negative classes. A model with a perfect AUROC of 1.0 has flawless discriminatory ability. Conversely, a classification model is regarded as having poor clinical discrimination if its AUROC falls below 0.8.[27] Accuracy measures the proportion of correct predictions a model makes out of all the prediction is makes. Sensitivity, also called recall or true positive rate, measures the proportion of actual positive cases that a model correctly identifies. Specificity, on the other hand, measures the proportion of actual negative cases. The F1-score is the harmonic mean of precision and recall, where precision, related to specificity, is the model's ability to identify positive cases out of all positive predictions made. The F1-score effectively balances the trade-off between precision and recall, making it a reliable and informative evaluation metric, particularly for imbalanced datasets.

***Subgroup Analysis***

After the final model was identified, subgroup analysis was performed to determine if there were variations in model performance among different race/ethnicity subgroups. Baseline prevalence of MASLD was first calculated for each subgroup to distinguish the difference between underlying disease prevalence from model bias. Then, model predictions were stratified by race/ethnicity categories, and standard model performance metrics were computed for each subgroup. Finally, pairwise *P*-values were completed to determine if a statistically significant difference exists between subgroups for each model performance metric.

***Fairness Postprocessing***

Disparities were quantified using differences in true positive rates across subgroups, consistent with the equal opportunity criterion, which aims to equalize sensitivity for all groups. To further investigate disparities relative to baseline prevalence, the positive predictive values and the negative predictive values for each subgroup were computed. A positive predictive value (PPV) is the ratio of true positives to the total number of positives while a negative predictive value (NPV) is the ratio of true negatives to the total number of negatives.[28] PPV and NPV account for subgroup-specific prevalence, highlighting whether over- or under-prediction in each subgroup was influenced by baseline disease prevalence.

To mitigate disparities attributed to model bias, equal opportunity postprocessing was applied with the ThresholdOptimizer from the Fairlearn package[29] which determines subgroup-specific decision thresholds. Subgroup model performance metrics were then recomputed, and statistical significance of differences in subgroup performance was evaluated using two proportion z-tests and bootstrap resampling.

## Results

***Study Population and Characteristics***

After implementing the inclusion/exclusion criteria in TriNetX with ICD-10-CM codes and in Spark with ICD-9-CM codes, there were 773,022 MASLD patients and 7,038,422 non-MASLD patients. There were no duplicates in the data. After splitting the data into training, validating, and testing, there were 541,638 patients in the MASLD training dataset, 115,600 in the validating dataset, and 115,784 in the testing dataset. In the non-MASLD cohort, there were 4,926,685 patients in the training dataset, 1,055,448 in the validating dataset, and 1,056,289 in the testing dataset. After removing patients with missing values for the selected features, there were 260,753 patients in the training dataset (MASLD=29,753; non-MASLD=231,000), 55,134 patients in the validating dataset (MASLD=6,287; non-MASLD=48,847), and 56,143 patients in the testing dataset (MASLD=6,297, non-MASLD=49,846).

Table 1 presents the demographic, clinical, and laboratory characteristics of individuals with and without MASLD before propensity score matching was deployed. Overall, these results highlight that individuals with MASLD demonstrate a distinct clinical profile characterized by greater metabolic burden, liver enzyme elevations, and features consistent with metabolic dysregulation. Further discussion can be found in the SMA Study Population Characteristics Discussion.

**Table 1. Baseline characteristics of the MASLD and Non-MASLD training cohorts: before matching.[a]**

| Characteristics | MASLD (N=29753) | Non-MASLD (N=231000) | *P*-value |
|---|---|---|---|
| **Sex** | | | |
| Male, n (%) | 12593 (42.33) | 95780 (41.46) | 0.005 |
| Female, n (%) | 17146 (57.63) | 134801 (58.36) | 0.02 |
| Unknown, n (%) | 14 (0.05) | 419 (0.18) | 1.0 |
| **Age** | | | |
| 18-34, n (%) | 3123 (10.5) | 41172 (17.82) | <0.001 |
| 35-49, n (%) | 7989 (26.85) | 65962 (28.55) | <0.001 |
| 50-64, n (%) | 10945 (36.79) | 62565 (27.08) | <0.001 |
| 65 or greater, n (%) | 7696 (25.87) | 61301 (26.54) | 0.01 |
| **Race/ethnicity** | | | |
| Non-Hispanic White, n (%) | 17927 (60.25) | 137023 (59.32) | 0.002 |
| Non-Hispanic Asian, n (%) | 2053 (6.9) | 18601 (8.05) | <0.001 |
| Non-Hispanic Black, n (%) | 2826 (9.5) | 37475 (16.22) | <0.001 |
| Non-Hispanic Other, n (%) | 1291 (4.34) | 8636 (3.74) | <0.001 |
| Hispanic, n (%) | 5656 (19.01) | 29265 (12.67) | <0.001 |
| T2DM, n (%) | 10171 (34.19) | 30817 (13.34) | <0.001 |
| Hypertension, n (%) | 17003 (57.15) | 85484 (37.01) | <0.001 |
| Smoking, n (%) | 3562 (11.97) | 18236 (7.89) | <0.001 |
| BMI, mean (SD) | 33.63 (6.54) | 29.04 (6.38) | <0.001 |
| TG, mean (SD) | 157.13 (88.22) | 114.20 (63.97) | <0.001 |
| ALT, mean (SD) | 47.25 (49.18) | 22.58 (29.41) | <0.001 |
| AST, mean (SD) | 35.25 (46.67) | 21.56 (33.55) | <0.001 |
| ALP, mean (SD) | 82.09 (36.75) | 71.37 (24.82) | <0.001 |
| BUN, mean (SD) | 13.99 (4.92) | 13.93 (4.81) | 0.05 |
| Cr, mean (SD) | 1.08 (2.71) | 1.21 (2.89) | <0.001 |
| BIL, mean (SD) | 0.60 (0.49) | 0.56 (0.31) | <0.001 |
| ALB, mean (SD) | 4.32 (0.39) | 4.42 (0.35) | <0.001 |
| TP, mean (SD) | 7.23 (0.51) | 7.15 (0.46) | <0.001 |
| FPG, mean (SD) | 110.82 (40.08) | 97.89 (29.18) | <0.001 |
| LDL, mean (SD) | 109.44 (36.13) | 110.27 (34.32) | 0.006 |
| HDL, mean (SD) | 47.36 (14.51) | 56.08 (16.52) | <0.001 |

[a]Abbreviations: MASLD, metabolic dysfunction-associated steatohepatitis; N, sample size; T2DM, type 2 diabetes mellitus; BMI, body mass index; TG, triglycerides; ALT, alanine transaminase; AST, aspartate aminotransferase; ALP, alkaline phosphatase; BUN, blood urea nitrogen; Cr, creatinine; BIL, bilirubin; ALB, albumin; TP, total protein; FPG, fasting plasma glucose; LDL, low density lipoprotein cholesterol; HDL, high density lipoprotein cholesterol.

***Propensity Score Matching***

Propensity score matching is only performed on the training cohort where the initial training set size is 260,753 patients. After performing propensity score matching (matching on sex and age), there were 29,753 MASLD patients and 29,739 non-MASLD patients left—a total sample size of 59,492 in the training cohort, and results are shown in Tables 1-2.

**Table 2. Baseline characteristics of the MASLD and Non-MASLD training cohorts: after matching.[a,b]**

| Characteristics | MASLD (N=29753) | Non-MASLD (N=231000) | *P*-value |
|---|---|---|---|
| **Sex[b]** | | | |
| Male, n (%) | 12593 (42.35) | 12593 (42.35) | 0.97 |
| Female, n (%) | 17146 (57.65) | 17146 (57.65) | 0.95 |
| Unknown, n (%) | 14 (0.05) | 0 | 1.0 |
| **Age[b]** | | | |
| 18-34, n (%) | 3123 (10.50) | 3119 (10.49) | 0.98 |
| 35-49, n (%) | 7989 (26.85) | 7988 (26.86) | 0.99 |
| 50-64, n (%) | 10945 (36.79) | 10941 (36.79) | 1.0 |
| 65 or greater, n (%) | 7696 (25.87) | 7691 (25.86) | 1.0 |
| **Race/ethnicity** | | | |
| Non-Hispanic White, n (%) | 17927 (60.25) | 17762 (59.73) | 0.19 |
| Non-Hispanic Asian, n (%) | 2053 (6.90) | 2350 (7.90) | <0.001 |
| Non-Hispanic Black, n (%) | 2826 (9.50) | 4944 (16.62) | <0.001 |
| Non-Hispanic Other, n (%) | 1291 (4.34) | 1104 (3.71) | <0.001 |
| Hispanic, n (%) | 5656 (19.01) | 3579 (12.03) | <0.001 |
| T2DM, n (%) | 10172 (34.19) | 4238 (14.25) | <0.001 |
| Hypertension, n (%) | 17003 (57.15) | 11833 (39.79) | <0.001 |
| Smoking, n (%) | 3562 (11.97) | 2512 (8.45) | <0.001 |
| BMI, mean (SD) | 33.63 (6.54) | 29.21 (6.34) | <0.001 |
| TG, mean (SD) | 157.13 (88.22) | 116.60 (64.53) | <0.001 |
| ALT, mean (SD) | 72.25 (49.18) | 22.81 (20.14) | <0.001 |
| AST, mean (SD) | 32.25 (46.67) | 21.60 (13.96) | <0.001 |
| ALP, mean (SD) | 82.09 (36.75) | 71.81 (24.99) | <0.001 |
| BUN, mean (SD) | 13.99 (4.92) | 14.13 (4.85) | <0.001 |
| Cr, mean (SD) | 1.08 (2.71) | 122 (2.95) | <0.001 |
| BIL, mean (SD) | 0.60 (0.49) | 0.57 (0.30) | 0.01 |

| ALB, mean (SD) | 4.32 (0.39) | 4.41 (0.35) | <0.001 |
|---|---|---|---|
| TP, mean (SD) | 7.23 (0.51) | 7.14 (0.46) | <0.001 |
| FPG, mean (SD) | 110.82 (40.08) | 98.61 (29.51) | <0.001 |
| LDL, mean (SD) | 109.44 (36.13) | 112.17 (34.64) | <0.001 |
| HDL, mean (SD) | 47.36 (14.51) | 56.32 (16.86) | <0.001 |

[a]Matched features are indicated by an asterisk.
[b]Abbreviations: MASLD, metabolic dysfunction-associated steatohepatitis; N, sample size; T2DM, type 2 diabetes mellitus; BMI, body mass index; TG, triglycerides; ALT, alanine transaminase; AST, aspartate aminotransferase; ALP, alkaline phosphatase; BUN, blood urea nitrogen; Cr, creatinine; BIL, bilirubin; ALB, albumin; TP, total protein; FPG, fasting plasma glucose; LDL, low density lipoprotein cholesterol; HDL, high density lipoprotein cholesterol.

***Validation Strategy***
Adjusting the proportion of MASLD to non-MASLD patients to be 1:3 resulted in 25,148 patients in the validation dataset (MASLD=6,287, non-MASLD=15,861) and 25,188 patients in the testing dataset (MASLD=6,297, non-MASLD=18,891). As mentioned before, this ensures modeling is tested under real epidemiological conditions.

***Modeling Results***
Table 3 summarizes the performance of each model on the testing set in terms of AUROC, accuracy, sensitivity, specificity, and F1 score, and SMA Table S5 displays each model's final hyperparameters. When using all features, the neural network and XGBoost models achieved the highest AUROC (0.850), followed closely by random forest (0.847) and logistic regression (0.843). In general, AUROC decreased as the number of input features was reduced; however, the decline in both AUROC and accuracy from the full-feature models to those using only the top 10 features was negligible.

**Table 3. Summary of model performances on testing data.[a]**

| **Model** | **AUROC** | **Accuracy (%)** | **Sensitivity (%)** | **Specificity (%)** | **F1** |
|---|---|---|---|---|---|
| LR (all features) | 0.843 | 78.0 | 72.9 | 79.6 | 0.623 |
| LR (top 10 features) | 0.840 | 77.6 | 72.3 | 79.3 | 0.617 |
| LR (top 5 features) | 0.829 | 76.5 | 72.7 | 77.8 | 0.608 |
| LR (top 3 features) | 0.812 | 76.1 | 67.8 | 78.9 | 0.587 |
| RF (all features) | 0.847 | 76.8 | 76.0 | 77.1 | 0.621 |
| RF (top 10 features) | 0.843 | 76.9 | 74.2 | 77.9 | 0.617 |
| RF (top 5 features) | 0.823 | 75.2 | 72.1 | 76.2 | 0.592 |
| RF (top 3 features) | 0.806 | 76.0 | 64.8 | 79.7 | 0.574 |
| XGB (all features) | 0.850 | 77.4 | 75.9 | 77.9 | 0.627 |
| XGB (top 10 features) | 0.847 | 77.2 | 75.4 | 77.8 | 0.623 |
| XGB (top 5 features) | 0.829 | 75.7 | 73.2 | 76.5 | 0.601 |

| XGB (top 3 features) | 0.816 | 74.5 | 72.0 | 75.4 | 0.586 |
|---|---|---|---|---|---|
| NN (all features) | 0.850 | 76.8 | 76.5 | 76.9 | 0.622 |
| NN (top 10 features) | 0.845 | 76.7 | 75.6 | 77.1 | 0.619 |
| NN (top 5 features) | 0.830 | 75.5 | 74.5 | 75.8 | 0.603 |
| NN (top 3 features) | 0.817 | 74.5 | 72.3 | 75.3 | 0.586 |

[a]Abbreviations: AUROC, area under receiver operating curve; LR, logistic regression; RF, random forest; XGB, extreme gradient boosting; NN, neural network.

Among the four model types, XGBoost with all features achieved the highest F1-score (0.627), followed closely by logistic regression (0.623), the neural network (0.622), and random forest (0.621). In terms of sensitivity, the neural network with all features performed best (76.5%), followed by random forest (76.0%) and XGBoost (75.9%). Logistic regression achieved the highest specificity (79.6%) and competitive F1 performance. Overall, all models demonstrated robust performance, particularly when using all or the top 10 features. However, due to its balance of interpretability, simplicity, and competitive performance, the logistic regression model with the top 10 SHAP features is a practical and effective choice for clinical implementation. The resulting LASSO logistic regression equation is given below:

$$\text{log_odds} = 0.6106 + 0.5583*\text{BMI} + 0.2036*\text{TG} + 1.5915*\text{ALT} + 0.5375*\text{AST} + -0.4076*\text{HDL} + -0.9625*\text{sex} + 0.8242*\text{T2DM} + 0.4840*\text{hypertension} + -0.3104*\text{Non-Hispanic White} + -1.0292*\text{Non-Hispanic Black} + -0.0885*\text{Non-Hispanic Asian} + -0.2108*\text{Non-Hispanic Other}$$

$$\text{Probability of having MASLD} = 1 / (1 + \exp(-\text{log_odds}))$$

The SHAP analysis shown in Figure 2 revealed consistent high-importance features across all models, including alanine aminotransferase (ALT), body mass index (BMI), sex, aspartate aminotransferase (AST), type 2 diabetes mellitus (T2DM), hypertension, high density lipoprotein (HDL), and race/ethnicity. These features were consistently ranked as top contributors to model performance and were used to construct reduced-feature models.

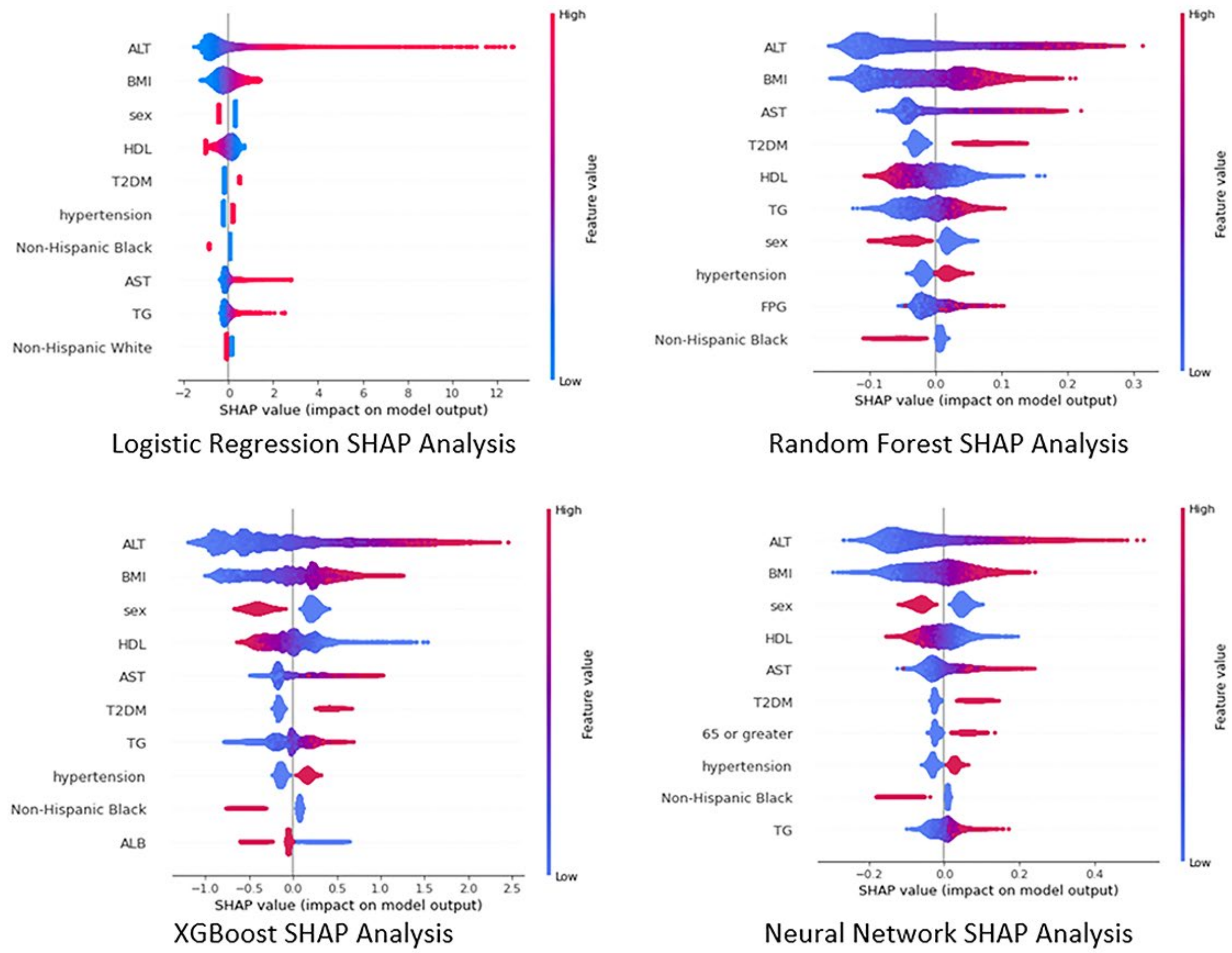

*Figure 2. SHAP Analysis Results. Abbreviations: ALB, albumin; ALT, alanine transaminase; AST, aspartate aminotransferase; BMI, body mass index; FPG, fasting plasma glucose; HDL, high density lipoprotein cholesterol; SHAP, shapley additive explanations; T2DM, type 2 diabetes mellitus; TG, triglycerides; XGBoost, extreme gradient boosting.*

***Subgroup Analysis***

First, the baseline prevalence of MASLD with each race/ethnicity subgroup in the testing data was computed. Prevalence is defined as the proportion of the subgroup who was diagnosed with MASLD. The results, shown in SMA Figure S1, indicate a slight difference in prevalence among race/ethnicity groups. This indicates that some variation in model performance across subgroups could reflect the true differences in disease prevalence.

Subgroup comparisons across model performance metrics are summarized in SMA Table S6. Pairwise *P*-values for each subgroup pair and a corresponding metric were computed. It is notable that AUROC does not have many significant differences between subgroups except Hispanic and non-Hispanic Black subgroups (*P*-value=0.02). Accuracy and sensitivity differ widely and significantly across many subgroup pairs, especially involving the Hispanic and non-Hispanic Black subgroups. Specificity differs in nearly all subgroup pairs. F1-Score comparisons show very few significant differences; the two differences that appear involve the Hispanic subgroup. Overall, AUROC and F1-score show fewer differences than accuracy, sensitivity, and specificity, which may point to threshold-dependent performance rather than overall model discrimination.

***Fairness Investigation and Analysis***

Before postprocessing, PPV roughly follows subgroup prevalence, indicating that the model's positive predictive value scales with disease frequency, disparities in the true positive rates (TPR) and the false positive rates (FPR) reveal true performance differences. Specifically, lower TPR in non-Hispanic Black patients indicates under-diagnosis, suggesting the model is biased in its detection despite PPV appearing consistent with prevalence. Therefore, fairness postprocessing is needed.

To select the fairness postprocessing method, we compared, equalized odds and equal opportunity whose results are presented in Figure 3. Equalized odds ensure TPR and FPR are equal across all subgroups while equal opportunity only focuses on TPR being equal across all subgroups[30]. After applying equalized odds postprocessing, TPR and FPR disparities were effectively eliminated, but overall sensitivity decreased substantially (eg, Non-Hispanic Black TPR dropped from 0.47 to 0.39), and NPV declined slightly. Equal opportunity achieved similar fairness in sensitivity (~0.41) while having some FPR variation (0.047-0.074). PPV and NPV were generally higher under the equal opportunity postprocessing meaning positive predictions were more clinically reliable and negative predictions remained trustworthy.

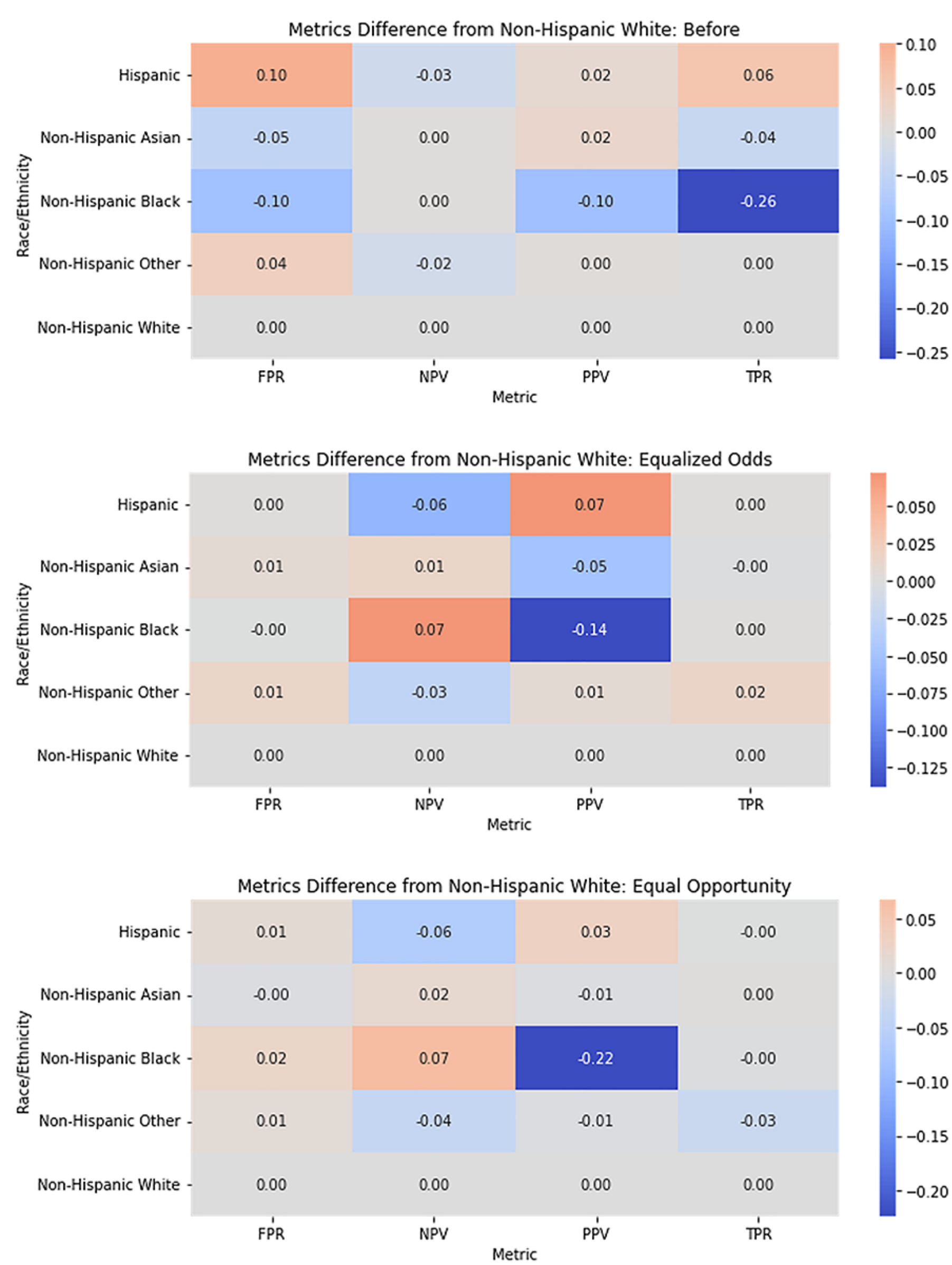

*Figure 3. Fairness Metrics Before and After Equalized Odds and Equal Opportunity Postprocessing. Abbreviations: FPR, false positive rate; NPV, negative predictive value; PPV, positive predictive value; TPR, true positive rate*

The primary goal of MASLD screening is to ensure positive cases are identified equitably across subgroups, while maintaining acceptable predictive reliability. Equal opportunity provides a more balanced trade-off between fairness and clinical utility; therefore, equal opportunity was selected as the postprocessing method.

***Fairness Postprocessing to Achieve Equal Opportunity Across Subgroups***

Now that the fairness postprocessing method has been selected, its impact on the LASSO logistic regression model with the top 10 SHAP values is evaluated using the test set. Before postprocessing, several subgroup comparisons show statistically significant differences in AUROC, accuracy, sensitivity, specificity, and F-1 score between subgroups. The results, in SMA Table S7, highlight the baseline disparities before applying the equal opportunity fairness postprocessing. SMA Table S8 displays the result after postprocessing. With higher *P*-values and fewer statistically significant pairwise differences after postprocessing, we observe that many disparities were reduced. Sensitivity became more balanced across subgroups, which is consistent with the constraint enforced by equal opportunity.

Applying the equal opportunity fairness method substantially altered both subgroup-level and overall model performance. At the subgroup level, TPR decreased for all race/ethnicity groups with the largest reductions observed for the Hispanic (-0.386) and the non-Hispanic White (-0.332) subgroups, while FPR also declined across all subgroups with the largest reductions being the Hispanic (-0.249) and non-Hispanic other (-0.204) subgroups, shown in Table 4. Positive predictive value improved for all groups whereas negative predictive value decreased for all groups (SMA Table S9).

**Table 4. True positive rate and false positive rate before and after applying fairness method.[a]**

| Subgroup | TPR Before | TPR After | Change in TPR | FPR Before | FPR After | Change in FPR |
|---|---|---|---|---|---|---|
| Non-Hispanic White | 0.727 | 0.395 | ↓ 0.332 | 0.209 | 0.048 | ↓ 0.161 |
| Non-Hispanic Asian | 0.706 | 0.456 | ↓ 0.250 | 0.152 | 0.044 | ↓ 0.108 |
| Non-Hispanic Black | 0.475 | 0.401 | ↓ 0.074 | 0.118 | 0.083 | ↓ 0.035 |
| Non-Hispanic Other | 0.767 | 0.455 | ↓ 0.312 | 0.271 | 0.067 | ↓ 0.204 |
| Hispanic | 0.823 | 0.437 | ↓ 0.386 | 0.322 | 0.073 | ↓ 0.249 |

[a]Abbreviations: TPR, true positive rate; FPR, false positive rate.

Accuracy of the overall model increased from 0.723 to 0.408, specificity increased substantially from 0.793 to 0.942, sensitivity decreased from 0.723 to 0.408, and F1-score decreased moderately from 0.617 to 0.516 as shown in Table 5 and SMA Table S10. McNemar's test, used to determine whether there is a significant difference in performance between two classifiers on the same dataset, also confirmed that the differences between pre- and postprocessing predictions were statistically significant ($\chi^2$=151.5141, p=0.0000) indicating that the fairness postprocessing meaningfully altered model decisions. The cost of fairness is evident in these results: although equal opportunity improved equity in TPR across subgroups, it came at the expense of lower overall sensitivity and moderate declines in NPV and F1-score.

***Comparison with Previous Studies***

The proposed LASSO logistic regression model using the top 10 SHAP features compares favorably with prior studies conducted on similar cohorts. Atsawarungruangkit et al (2021) proposed two interpretable models for potential clinical use: an ensemble of RUS-boosted trees with 30 features (AUROC=0.79, F1-score=0.56) and a simplified coarse decision tree using only two features (AUROC=0.56) [31]. Zhu et al (2025) developed a logistic regression model with eight features, reporting an AUROC of 0.806 (F1-score not reported) [32]. Similarly, Nouredin et al (2022) used ten features in their model and reported an AUROC of 0.830, concluding that logistic regression was the most suitable method for clinical translation [33]. Collectively, these studies reinforce the importance of selecting a simple, interpretable model and further support the strong performance and practical utility of the proposed LASSO logistic regression model.

**Table 5. Model performance comparison with MASLD prediction models of similar study populations.[a]**

| Reference | Study Population | Train/ Validate / Test (N) | Model | AUROC | ACC (%) | SEN (%) | SPE (%) | F1(%) |
|---|---|---|---|---|---|---|---|---|
| Atsawarungruangkit et al, 2021[31] | US mixed population (NHANES) | 2265/ 0/ 970 | Ensemble of RUS boosted trees | 0.79 | 71.1 | 72.7 | 70.6 | 0.56 |
| Atsawarungruangkit et al, 2021[31] | US mixed population (NHANES) | 2265/ 0/ 970 | Coarse trees | 0.72 | 74.9 | 24.5 | 92.0 | 0.33 |
| Zhu et al, 2025[32] | Chinese hospital; external validation with US mixed population (NHANES) | 7003/ 2002/ 1002 | Logistic regression | 0.81 | 72.8 | 74.9 | 71.3 | 0.701 |
| Noureddin et al., 2022[33] | US mixed population (NHANES) | 2874/ 0/ 957 | Logistic regression | 0.83 | 78 | 55 | 89 | N/A |
| Our Model | US mixed population (TriNetX) | 59492/ 24198/ 25188 | LASSO logistic regression | 0.84 | 77.6 | 72.3 | 79.3 | 0.617 |
| | | | LASSO logistic regression with fairness postprocessing | N/A[b] | 80.8 | 40.7 | 94.2 | 0.515 |

[a]Abbreviations: MASLD, metabolic dysfunction-associated steatohepatitis; N, sample size; AUROC, area under the receiver operating curve; ACC, accuracy; SEN, sensitivity; SPE, specificity; U.S., United States; NHANES, national health and nutrition examination survey; RUS, random undersampling; LASSO, least absolute shrinkage and selection operator; N/A, value not available.

[b]Value is not available since the Fairlearn Threshold Optimizer uses discrete, optimized thresholds rather than the full probability spectrum.

### *Online Prediction Tool: MASLD Static EHR Risk Prediction (MASER)*

An online MASER tool[34] was developed to enable application of the LASSO logistic regression model directly to EHR-derived data (refer to SMA Figures S2-S4). Rather than a traditional risk calculator, it provides a deployment interface for real or sample patient data, facilitating exploration of its predictive performance in a clinical implementation context.

## Discussion

This study developed and evaluated multiple machine learning models for predicting MASLD using EHR data. Among the four approaches evaluated, namely LASSO logistic regression, random forest, XGBoost, and a neural network, XGBoost achieved the highest overall AUROC (0.85) and F1-score (0.627) when using all features. The neural network demonstrated the best sensitivity (76.5%), while random forest had the highest specificity (79.7%) with LASSO logistic regression slightly behind. LASSO logistic regression achieved the best accuracy (78%) with all features. Notably, models using only the top 10 SHAP-selected features performed comparably to those using the full feature set, highlighting the potential for efficient, interpretable prediction with reduced input complexity.

When applied to the LASSO logistic regression model, the equal opportunity approach substantially reduced disparities based on true positive rates across racial and ethnic subgroups while maintaining acceptable overall accuracy, though this improvement came at the cost of lower sensitivity and moderate reductions in F1-score and NPV. This combination of LASSO logistic regression with SHAP-based feature selection and fairness-aware postprocessing offers a robust framework that supports both equitable and clinically meaningful decision-making.

Although the neural network and XGBoost models demonstrated slightly better overall predictive performance, LASSO logistic regression remains the preferred method for clinical settings due to its interpretability and explainability. This transparency is especially important in clinical decisions like MASLD diagnosis, where explainability can influence provider trust, patient communication, and ethical implementation. Furthermore, the LASSO logistic regression modeled with the top 10 SHAP features outperformed comparable approaches in the literature, achieving the highest AUROC, and demonstrated comparable or superior performance in terms of accuracy, sensitivity, specificity, and F1-score. Therefore, the LASSO logistic regression model using the top 10 SHAP-selected features is recommended.

Furthermore, we restricted our model to features that are routinely collected during standard primary care visits. Several previously published models for MASLD prediction have incorporated additional anthropometric measures such as waist circumference, which, although highly informative, are not consistently collected in routine practice and would add to the clinical burden. In contract, our model relies exclusively on features that are part of standard outpatient workflows. This design choice enhances the practicality and scalability of the model for real-world implementation while still maintaining strong predictive performance.

### *Limitations*

Several limitations should be considered when interpreting our findings. First, TriNetX data is primarily from teaching hospitals which may have a different population distribution from non-teaching hospitals or from the general US population. Additionally, EHR data is only available for those patients who utilize healthcare. Therefore, the dataset used in this study may not be fully representative of all demographic groups or all healthcare settings and may demonstrate misclassification bias. It was observed that although many patients in the EHR met the clinical criteria for MASLD, they lacked a corresponding MASLD ICD-10-CM code, indicating underdiagnosis, which may have impacted model performance.[35] Second, while the equal opportunity method reduced differences in sensitivity across the race/ethnicity subgroups, it came at the cost of reduced overall sensitivity and declines in F1-score and NPV. Third, PPV and NPV are influenced by baseline disease prevalence, which complicates interpretation across subgroups, and other attributes such as insurance type or socioeconomic status that were not included in the analysis. Finally, the LASSO logistic regression model may not capture non-linear or longitudinal relationships in EHR data which could limit overall performance. These factors highlight the need for cautious interpretation and suggest that future work is needed to enhance both fairness and clinical utility.

### *Clinical Implications*

The model has been made publicly available online to facilitate clinical adoption and translational use. The web-based MASLD Static EHR Risk Prediction (MASER) interface allows users to input patient-level data and obtain real-time MASLD risk predictions, making it accessible to both clinicians and researchers without requiring advanced computational expertise. Because all MASER features are derived directly from routinely collected EHR variables and utilize standard LOINC and ICD-10-CM codes, the model can be seamlessly integrated into existing clinical workflows or embedded within health information systems. This design minimizes additional data entry burden and ensures compatibility across diverse healthcare settings. By leveraging information already available in the patient records, MASER enables scalable, low-cost implementation and supports early identification of at-risk individuals. Broad application of this model may enhance MASLD detection rates, promote timely intervention, and contribute to improved population-level liver health outcomes.

### *Conclusions*

This study presents MASER, an interpretable and equitable machine learning model for MASLD prediction from EHR data. By combining strong predictive performance with transparency and fairness, MASER bridges the gap between algorithmic development and real-world clinical implementation, offering a pathway toward earlier diagnosis and improved patient outcomes.

## Acknowledgments
An artificial intelligence–based language tool, OpenAI ChatGPT, was used for generating part of the manuscript text and for grammar and style. All scientific content was generated, verified, and approved by the authors.

## Funding
This work was supported by the National Institute of Standards and Technology (ror.org/05xpvk416). Mary Ogidigben (An) was supported through PREP agreement no. 60NANB19D107 between NIST and the Pennsylvania State University (ror.org/04p491231). The content is solely the responsibility of the authors and does not necessarily represent the official views of NIST.

Use of the TriNetX dataset was supported by the National Center for Advancing Translational Sciences, National Institutes of Health, through Grant UL1 TR002014. The content is solely the responsibility of the authors and does not necessarily represent the official views of the NIH.

## Conflicts of Interest
Jonathan G. Stine receives salary support from Astra Zeneca, research grant support from Astra Zeneca, Galectin, Kowa, Novo Nordisk, Regeneron, Zydus and is an advisory board member for Madrigal. All other authors declare no conflicts of interest.

## Data Availability
The TriNetX data cannot be shared. However, the supplementary multimedia appendix includes the inclusion and exclusion criteria and their accompanying ICD-10-CM codes used to extract the TriNetX data. Additionally, the code used for preprocessing and modeling and the MASER online prediction tool has been made available at: https://github.com/mary-elena-an/MASLD-EHR-Prediction.

## Authors' Contributions
Mary E An: conceptualization, data curation, formal analysis, methodology, software, visualization, original draft, writing – review and editing; Paul M Griffin: conceptualization, methodology, project administration, writing – original draft, writing – review and editing; Jonathan G Stine: conceptualization, methodology, writing – review and editing; Balakrishnan S Ramakrishna: conceptualization, writing – review and editing; Soundar RT Kumara: conceptualization, funding acquisition, methodology, resources, supervision, writing – original draft, writing – review and editing.

## Abbreviations
ACC, accuracy; ALB, albumin; ALP, alkaline phosphatase; ALT, alanine transaminase; AST, aspartate aminotransferase; BIL, bilirubin; BMI, body mass index; BUN, blood urea nitrogen; Cr, creatinine; EHR, electronic health record; FLI, fatty liver index; FPG, fasting plasma glucose; FPR, false positive rate; HDL, high density lipoprotein cholesterol; HIPAA, health insurance portability and accountability act; HSI, hepatic steatosis index; ICD-9-CM, international classification of diseases, 9$^{th}$ revision, clinical medication; ICD-10-CM, international classification of diseases, 10$^{th}$ revision, clinical medication; ID, identification; IRB, internal review board; LASSO, least absolute shrinkage and selection operator; LDL, low density lipoprotein cholesterol; LR, logistic regression; MASER, metabolic dysfunction-associated steatotic liver disease electronic health record static risk prediction; MASH, metabolic dysfunction-associated steatohepatitis; MASLD, metabolic dysfunction-associated steatotic liver disease; N, sample size; N/A, value not available; NASH, non-alcoholic steatohepatitis; NHANES, national health and nutrition examination survey; NN, neural network; NPV, negative predictive value; PNPLA3 gene, patatin-like phospholipase domain-containing 3 gene; PPV, positive predictive value; RF, random forest; ROC, receiver operating characteristic; RUS-boosted trees, random undersampling boosting trees; SEN, sensitivity; SHAP, shapley additive explanations; SMA, supplementary multimedia appendix; SMOTETomek, synthetic minority over-sampling technique and tomek links; SPE, specificity; T2DM, type 2 diabetes mellitus; TG, triglycerides; TP, total protein; TPR, true positive rate; US, United States; XGB, extreme gradient boosting; XGBoost, extreme gradient boosting; ZJU, Zhejiang University Index

## Supplementary Multimedia Appendix File

Supplementary Multimedia Appendix 1 includes detailed cohort inclusion and exclusion criteria (SMA Tables S1–S4), study population characteristics, machine learning model hyperparameters and performance metrics (SMA Tables S5–S10), and figures illustrating MASLD prevalence and the MASER tool interface (SMA Figures S1–S4). These supplementary materials provide additional methodological detail and support reproducibility of the study.

**SMA, Table S1. Inclusion Criteria to Filter "NAFLD" Study Cohort**

| **Category** | **Inclusion Criteria** |
|---|---|
| Age | 18 year or older |
| Fatty Liver | ICD-10-CM Code: K76.0<br>ICD-9-CM Code: 5718 |
| Non-alcoholic steatohepatitis (NASH) | ICD-10-CM Code: K75.81<br>ICD-9-CM Code: 5733 |

**SMA, Table S2. Inclusion and Exclusion Criteria to Filter "Non-NAFLD" Study Cohort**

| **Category** | **Inclusion Criteria** |
|---|---|
| Age | 18 year or older |
| Adult Medical Examination | ICD-10-CM Code: Z00.00<br>ICD-9-CM Code: V70.0 |
| **Category** | **Exclusion Criteria** |
| Fatty Liver | ICD-10-CM Code: K76.0<br>ICD-9-CM Code: 5718 |
| Non-alcoholic steatohepatitis (NASH) | ICD-10-CM Code: K75.81<br>ICD-9-CM Code: 5733 |

**SMA, Table S3. Exclusion Criteria to Filter "NAFLD" and "Non-NAFLD" Study Cohorts**

| **Category** | **Subcategory** | **Exclusion Criteria** | **ICD-10-CM** |
|---|---|---|---|
| Alcohol | | Alcohol related disorders | F10 |
| Other causes of liver disease | Hepatitis | Autoimmune hepatitis | K75.4 |
| | | Chronis hepatitis | K73 |
| | | Toxic hepatitis | K71.6 |

| | | Viral hepatitis | B15-B19 |
|---|---|---|---|
| | Inherited metabolic disorders and other genetic diseases | A1a deficiency | E88.01 |
| | | Budd Chiari syndrome | I82.0 |
| | | Gaucher disease | E75.22 |
| | | Glycogen storage disease (GSD) | E74.00 |
| | | Hemochromatosis | E83.110 |
| | | Wilson's disease | E83.01 |
| | Autoimmune disorders | Primary biliary cholangitis | K74.3 |
| | | Primary sclerosing cholangitis | K83.01 |
| | Cardiovascular diseases affecting the liver | Arterial diseases | I70-I79 |
| | | Ischemia heart diseases | I20-I25 |
| | | Right sided heart failure | I50.81 |
| | Cancer | Bile duct cancer | C22 |
| | | Liver cancer | C22 |
| | Biliary stasis | Biliary stasis | K83.1 |
| Secondary causes of steatosis | | Long term use of Tamoxifen | Z79.810 |

**SMA, Extended Data Table. Exclusion Criteria to Filter "NAFLD" Study Cohort: ICD-9-CM Codes Corresponding to ICD-10-CM Codes**

Use the following link to access Extended Data Table – Exclusion Criteria: Extended Data Table - Exclusion Criteria.xlsx

**SMA, Table S4. ICD-10-CM, ICD-9-CM, LOINC codes, and capping values of Demographic and Clinical Features**

| Feature | Units | ICD-10-CM | ICD-9-CM | LOINC | Demographic |
|---|---|---|---|---|---|
| ALT | U/L | | | 1742-6 | |
| AST | U/L | | | 1920-8 | |
| BMI | Ratio | | | 39156-5 | |
| HDL | mg/dL | | | 2085-9 | |
| LDL | mg/dL | | | 13457-7 | |

| TG | mg/dL | | | 2571-8 | |
|---|---|---|---|---|---|
| BUN | mg/dL | | | 3094-0 | |
| Cr | mg/dL | | | 2160-0 | |
| BIL | mg/dL | | | 1975-2 | |
| ALB | g/dL | | | 1751-7 | |
| TP | g/dL | | | 2885-2 | |
| FPG | mg/dL | | | 2345-7 | |
| T2DM | | E11 | 250 | | |
| Hypertension | | I10,I11,I12,I13,<br>I15,I16,I1A | | | |
| Smoker (current) | | Z72.0<br>F17.2 | 305.1 | | |
| Sex | | | | | Male, Female |
| Race/Ethnicity | | | | | Non-Hispanic White, Non-Hispanic Asian, Non-Hispanic Black, Non-Hispanic Other, Hispanic |
| Age | | | | | 18-34, 35-49, 50-64, 65 or greater |

Abbreviations: ALT = alanine aminotransferase, AST = aspartate aminotransferase, BMI = body mass index, HDL = high density lipoprotein, LDL = low density lipoprotein, TG = triglycerides, BUN = blood urea nitrogen, Cr = creatinine, BIL = bilirubin, ALB = albumin, TP = total protein, FPG = fasting plasma glucose, T2DM = type 2 diabetes mellitus

**SMA, Study Population Characteristics Discussion**

Compared to the non-MASLD group (n=231,000), individuals with MASLD (n=29,753) were more often male (42.3% vs. 41.5%, $p = 0.0045$) and tended to be older, with a higher proportion aged 50–64 years (36.8% vs. 27.1%, $p < 0.001$) and fewer aged 18–34 years (10.5% vs. 17.8%, $p < 0.001$). In terms of race and ethnicity, MASLD was more prevalent among Hispanic (19.0% vs. 12.7%) and non-Hispanic White individuals (60.3% vs. 59.3%), while non-Hispanic Black (9.5% vs. 16.2%) and non-Hispanic Asian (6.9% vs. 8.1%) individuals were less represented ($p < 0.001$ for all comparisons).

Comorbid conditions were markedly more common in the MASLD cohort, including type 2 diabetes mellitus (34.2% vs. 13.3%, $p < 0.001$), hypertension (57.2% vs. 37.0%, $p < 0.001$), and smoking (12.0% vs. 7.9%, $p < 0.001$). Laboratory findings revealed significant metabolic and hepatic differences. Individuals with MASLD had higher mean BMI (33.63 ± 6.54 vs. 29.04 ± 6.38), triglycerides (157.13 ± 88.22 vs. 114.20 ± 63.97), fasting plasma glucose (110.82 ± 40.08 vs. 97.89 ± 29.18), and liver enzymes (ALT, AST, ALP; all $p < 0.001$). Conversely, MASLD participants exhibited lower HDL cholesterol (47.36 ± 14.51 vs.

56.08 ± 16.52) and serum albumin (4.32 ± 0.39 vs. 4.42 ± 0.35), while total protein and bilirubin were modestly higher. No significant difference was observed for blood urea nitrogen ($p$ = 0.0536).

**SMA, Table S5. Final Hyperparameters of Machine Learning Models**

| Model Name | Model Type | penalty | C | solver | max_iter | random_state |
|---|---|---|---|---|---|---|
| final_logistic_model | Logistic Regression | l1 | 0.1 | liblinear | 1000 | 42 |
| final_logistic_top10_model | Logistic Regression | l1 | 1 | liblinear | 1000 | 42 |
| final_logistic_top3_model | Logistic Regression | l1 | 1 | liblinear | 1000 | 42 |
| final_logistic_top5_model | Logistic Regression | l1 | 1 | liblinear | 1000 | 42 |

| Model Name | Model Type | n_estimators | max_depth | min_samples_split | min_samples_leaf | random_state |
|---|---|---|---|---|---|---|
| final_rf_model | Random Forest | 200 | 20 | 5 | 1 | 42 |
| final_rf_top10_model | Random Forest | 200 | 10 | 5 | 1 | 42 |
| final_rf_top3_model | Random Forest | 30 | 3 | 2 | 1 | 42 |
| final_rf_top5_model | Random Forest | 200 | 5 | 2 | 1 | 42 |

| Model Name | Model Type | n_estimators | max_depth | learning_rate | subsample | colsample_bytree | random_state |
|---|---|---|---|---|---|---|---|
| final_xgb_model | XGBClassifier | 300 | 3 | 0.1 | 0.7 | 1 | 42 |
| final_xgb_top10_model | XGBClassifier | 300 | 3 | 0.1 | 0.8 | 0.8 | 42 |
| final_xgb_top3_model | XGBClassifier | 200 | 6 | 0.01 | 0.7 | 0.7 | 42 |
| final_xgb_top5_model | XGBClassifier | 300 | 3 | 0.1 | 0.8 | 0.7 | 42 |

| **Model Name** | **Model Type** | **hidden_layer_sizes** | **activation** | **learning_rate_init** | **max_iter** | **random_state** |
|---|---|---|---|---|---|---|
| final_nn_model | NN - Keras Classifier | 160-96 | relu-relu | 0.000253592 | 20 | 42 |
| final_nn_top10_model | NN - Keras Classifier | 160-96 | relu-relu | 0.000253592 | 20 | 42 |
| final_nn_top3_model | NN - Keras Classifier | 160-96 | relu-relu | 0.000253592 | 20 | 42 |
| final_nn_top5_model | NN - Keras Classifier | 160-96 | relu-relu | 0.000253592 | 20 | 42 |

**SMA, Figure S1. MASLD Prevalence by Race/Ethnicity Group**

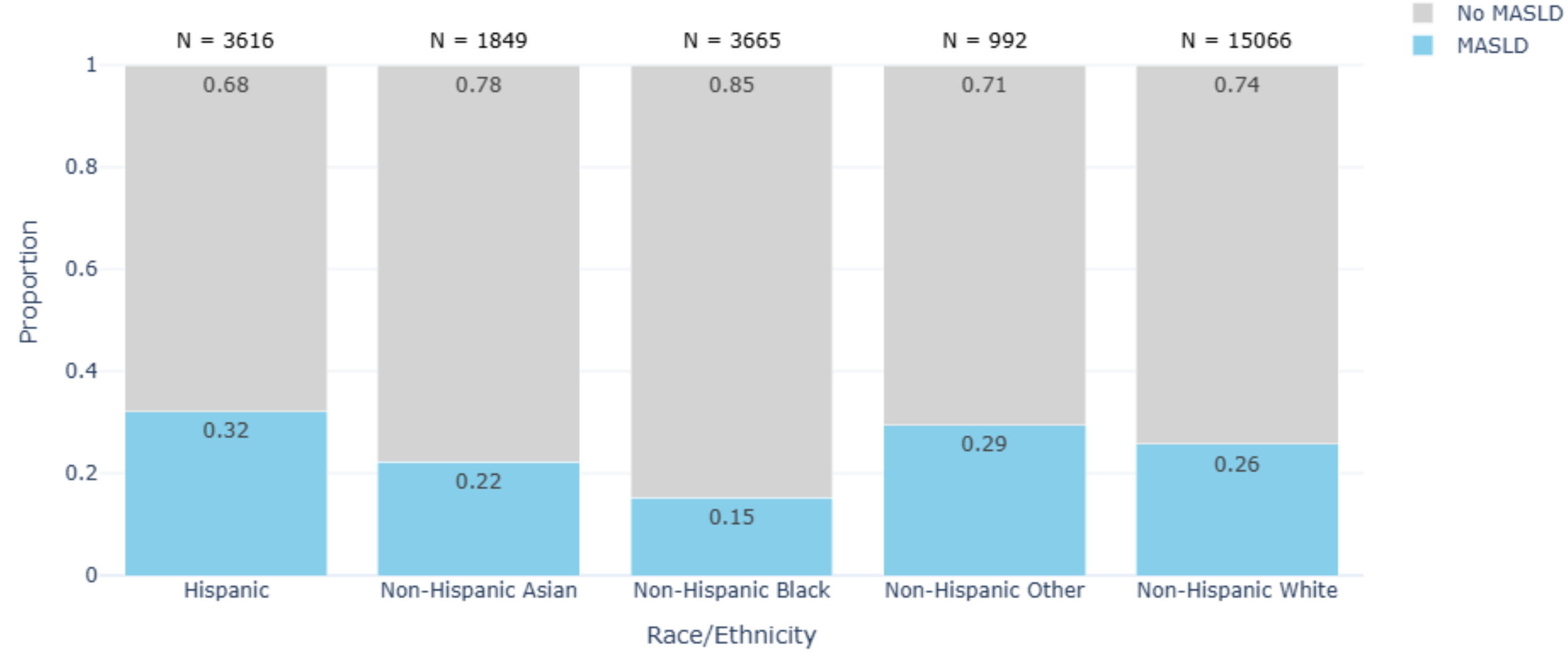

*Figure S1. MASLD Prevalence by Race/Ethnicity Subgroup*

**SMA, Table S6. Subgroup Analysis of LASSO Logistic Regression Model with Top 10 SHAP Features on Validating Dataset BEFORE Applying Fairness Method**

| Metric | Non-Hispanic White | Non-Hispanic Asian | Non-Hispanic Black | Non-Hispanic Other |
|---|---|---|---|---|
| **AUROC** | | | | |
| Non-Hispanic Asian | 0.883 | | | |
| Non-Hispanic Black | 1 | 1 | | |
| Non-Hispanic Other | 0.187 | 0.943 | 0.942 | |
| Hispanic | 0.987 | 0.993 | 0.02* | 0.576 |
| **Accuracy** | | | | |
| Non-Hispanic Asian | 0.003* | | | |
| Non-Hispanic Black | 2.12e-10* | 0.093 | | |
| Non-Hispanic Other | 0.028* | 0.000* | 3.41e-08* | |
| Hispanic | 7.22e-11* | 1.91e-11* | 3.09e-24* | 0.199 |
| **Sensitivity** | | | | |
| Non-Hispanic Asian | 0.073 | | | |
| Non-Hispanic Black | 8.91e-37* | 1.13e-12* | | |
| Non-Hispanic Other | 0.990 | 0.246 | 5.62e-13* | |
| Hispanic | 2.41e-05* | 1.58e-05* | 9.76e-43* | 0.027* |
| **Specificity** | | | | |
| Non-Hispanic Asian | 7.85e-06* | | | |
| Non-Hispanic Black | 1.19e-35* | 2.30e-06* | | |
| Non-Hispanic Other | 0.012* | 6.36e-07* | 6.78e-22* | |
| Hispanic | 2.56e-26* | 1.01e-25* | 9.02e-76* | 0.002* |
| **F1-Score** | | | | |
| Non-Hispanic Asian | 0.513 | | | |
| Non-Hispanic Black | 1 | 1 | | |
| Non-Hispanic Other | 0.551 | 0.501 | 1 | |
| Hispanic | 0* | 0.056 | 0* | 0.082 |

*Note: The values in the table represent p-values from pairwise comparisons between two race/ethnicity groups. A p-value < 0.05 indicates a statistically significant difference between the groups for the corresponding metric. P-values < 0.05 are indicated by an asterisk (*)*

**SMA, Table S7. Subgroup Analysis of LASSO Logistic Regression Model with Top 10 SHAP Features on Testing Dataset BEFORE Applying Equal Opportunity Fairness Method**

| Metric | Non-Hispanic White | Non-Hispanic Asian | Non-Hispanic Black | Non-Hispanic Other |
|---|---|---|---|---|
| **AUROC** | | | | |
| Non-Hispanic Asian | 0.997 | | | |
| Non-Hispanic Black | 0* | 1 | | |

| Non-Hispanic Other | 0.415 | 0.023* | 0.983 | |
|---|---|---|---|---|
| Hispanic | 0.077 | 1 | 0.006* | 0.703 |
| **Accuracy** | | | | |
| Non-Hispanic Asian | 4.86e-05* | | | |
| Non-Hispanic Black | 2.88e-09* | 0.748 | | |
| Non-Hispanic Other | 0.012* | 2.04e-06* | 2.18e-08* | |
| Hispanic | 2.35e-10* | 1.08e-13* | 4.93e-22* | 0.3344 |
| **Sensitivity** | | | | |
| Non-Hispanic Asian | 0.375 | | | |
| Non-Hispanic Black | 1.27e-33* | 5.92e-13* | | |
| Non-Hispanic Other | 0.135 | 0.073 | 2.73e-16* | |
| Hispanic | 3.80e-11* | 5.61e-07* | 4.41e-50* | 0.030* |
| **Specificity** | | | | |
| Non-Hispanic Asian | 5.28e-07* | | | |
| Non-Hispanic Black | 4.89e-30* | 0.002* | | |
| Non-Hispanic Other | 8.71e-05* | 5.22e-11* | 4.70e-25* | |
| Hispanic | 2.67e-33* | 3.24e-31* | 1.31e-76* | 0.011* |
| **F1-Score** | | | | |
| Non-Hispanic Asian | 0.635 | | | |
| Non-Hispanic Black | 0* | 1 | | |
| Non-Hispanic Other | 0.696 | 0.541 | 1 | |
| Hispanic | 0.997 | 0.102 | 0* | 0.153 |

*Note: The values in the table represent p-values from pairwise comparisons between two race/ethnicity groups. A p-value < 0.05 indicates a statistically significant difference between the groups for the corresponding metric. P-values < 0.05 are indicated by an asterisk (*)*

**SMA, Table S8. Subgroup Analysis of LASSO Logistic Regression Model with Top 10 SHAP Features on Testing Dataset AFTER Applying Equal Opportunity Fairness Method**

| Metric | Non-Hispanic White | Non-Hispanic Asian | Non-Hispanic Black | Non-Hispanic Other |
|---|---|---|---|---|
| **AUROC** | | | | |
| Non-Hispanic Asian | N/A | | | |
| Non-Hispanic Black | N/A | N/A | | |
| Non-Hispanic Other | N/A | N/A | N/A | |
| Hispanic | N/A | N/A | N/A | N/A |
| **Accuracy** | | | | |
| Non-Hispanic Asian | 1.04e-03* | | | |
| Non-Hispanic Black | 5.93e-05* | 0.790 | | |
| Non-Hispanic Other | 0.047* | 1.53e-04* | 6.10e-05* | |
| Hispanic | 4.70e-08* | 5.37e-10* | 1.18e-13* | 0.327 |
| **Sensitivity** | | | | |
| Non-Hispanic Asian | 0.052 | | | |
| Non-Hispanic Black | 0.764 | 0.182 | | |
| Non-Hispanic Other | 0.142 | 0.878 | 0.299 | |
| Hispanic | 8.43e-03* | 0.829 | 0.151 | 0.992 |
| **Specificity** | | | | |
| Non-Hispanic Asian | 0.754 | | | |
| Non-Hispanic Black | 6.83e-15* | 3.87e-06* | | |
| Non-Hispanic Other | 0.003* | 0.001* | 0.331 | |
| Hispanic | 4.80e-08* | 3.61e-04* | 0.218 | 0.850 |
| **F1-Score** | | | | |
| Non-Hispanic Asian | 0.079 | | | |
| Non-Hispanic Black | 1 | 0* | | |
| Non-Hispanic Other | 0.159 | 0.408 | 1 | |
| Hispanic | 0.002* | 0.561 | 0* | 0.427 |

*Note: The values in the table represent p-values from pairwise comparisons between two race/ethnicity groups. A p-value < 0.05 indicates a statistically significant difference between the groups for the corresponding metric. P-values < 0.05 are indicated by an asterisk (*)*

**SMA, Table S9. Positive Predictive Value (PPV) and Negative Predictive Value (NPV) Before and After Applying Fairness Method**

| Subgroup | PPV Before | PPV After | Change in PPV | NPV Before | NPV After | Change in NPV |
|---|---|---|---|---|---|---|
| Non-Hispanic White | 0.546 | 0.740 | ↑ 0.194 | 0.893 | 0.820 | ↓ 0.073 |
| Non-Hispanic Asian | 0.571 | 0.750 | ↑ 0.179 | 0.910 | 0.860 | ↓ 0.050 |
| Non-Hispanic Black | 0.419 | 0.464 | ↑ 0.045 | 0.903 | 0.895 | ↓ 0.008 |
| Non-Hispanic Other | 0.541 | 0.739 | ↑ 0.198 | 0.882 | 0.802 | ↓ 0.080 |
| Hispanic | 0.548 | 0.740 | ↑ 0.192 | 0.890 | 0.820 | ↓ 0.070 |

**SMA, Table S10. Model Performance Before and After Applying Fairness Method**

| Metric | Before | After | Change in Metric (After − Before) | Summary |
|---|---|---|---|---|
| Accuracy | 0.776 | 0.809 | +0.033 | Slightly improved |
| Sensitivity | 0.723 | 0.408 | −0.314 | Large decrease |
| Specificity | 0.793 | 0.942 | +0.149 | Moderate increase |
| F1-Score | 0.617 | 0.516 | −0.101 | Moderate decrease |

**SMA, Figure S2. MASER Tool – Not Filled Out**

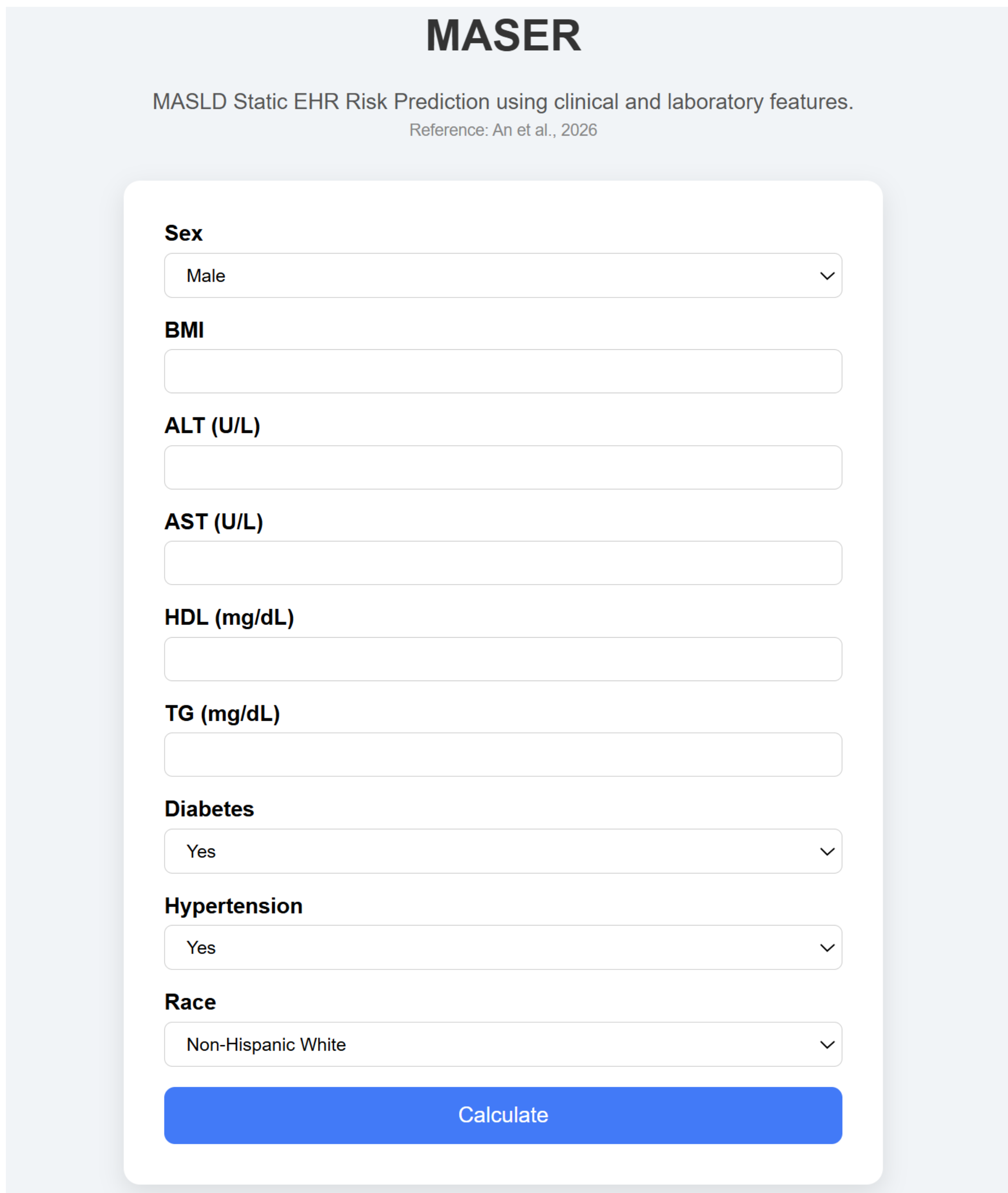

*Figure S2. MASER Tool – Not Filled Out*

**SMA, Figure S3. MASER Tool – Filled Out the Tool**

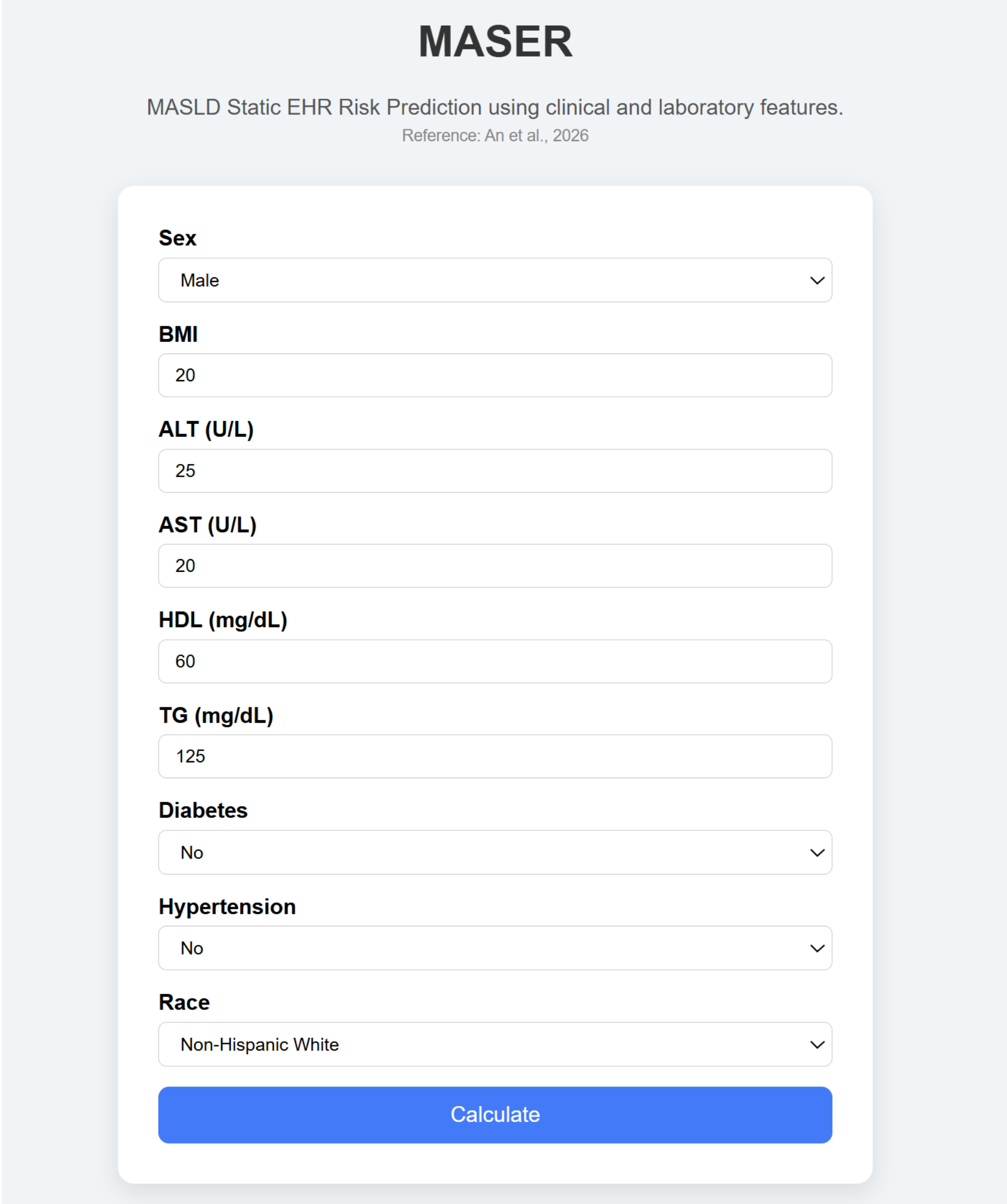

*Figure S3. MASER Tool – Filled Out the Tool*

**SMA, Figure S4. MASER Tool – Pressed "Calculate" and the Probability of Having MASLD or Not Having MASLD is Given**

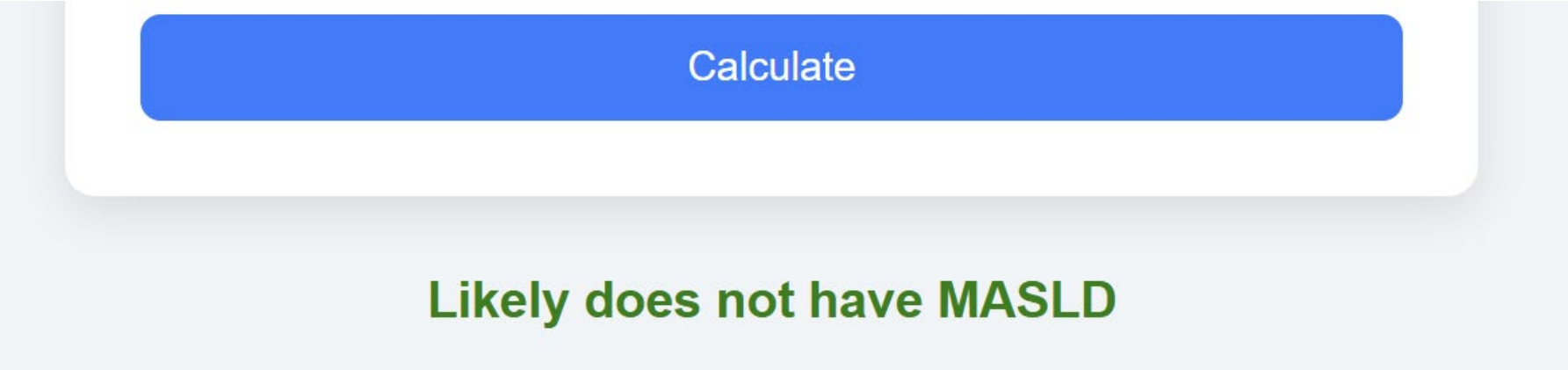

*Figure S4. MASER Tool – Pressed "Calculate" and the Probability of Having MASLD or Not Having MASLD is Given*